\documentclass[pdflatex,sn-mathphys-num]{sn-jnl}

\usepackage{graphicx,amsmath,amssymb}
\usepackage{booktabs}
\usepackage{tikz}
\usetikzlibrary{arrows.meta,positioning,shapes.geometric}
\usepackage{xcolor}
\usepackage{url}
\usepackage{enumitem}
\usepackage{multirow}
\newif\ifarxiv
\arxivtrue
\ifarxiv
\fi

\makeatletter
\gdef\orcidlogo{}%
\makeatother

\newcommand{\system}{\textsc{Viveka-Insight}}

\begin{document}

\title[A bilingual concept graph over Vivekananda's works]{\system{}: a
cross-lingual concept graph and citation-grounded retrieval resource over
the complete works of Swami Vivekananda in English and Bengali}

\author*[1]{\fnm{Tamal} \sur{Maharaj}}
\email{tamal@gm.rkmvu.ac.in}

\affil[1]{\orgname{Ramakrishna Mission Vivekananda Educational and
Research Institute}, \orgaddress{\city{Belur Math}, \state{West Bengal},
\country{India}}}

\abstract{
Classical spiritual and philosophical corpora pose three compounding
challenges for language resources: they exist in several languages
without parallel alignment, their vocabulary differs sharply from that of
contemporary readers, and any generated text over such culturally
sensitive material must be verifiably grounded in the source. We present
\system{}, a bilingual resource and accompanying open-source pipeline for
the works of Swami Vivekananda (1863--1902) --- the nine-volume English
\emph{Complete Works} and the ten-volume Bengali \emph{Vani o Rachana},
two related but non-parallel corpora totalling about 15 million
characters. The released resource has four layers: (i) a
structure-preserving parse of both corpora into a
volume\,$\to$\,chapter\,$\to$\,paragraph\,$\to$\,sentence hierarchy
(32{,}694 paragraphs, 168{,}842 sentences) with per-paragraph anchors
that deep-link back to the published editions; (ii) a \emph{cross-lingual
concept graph} of 8{,}362 language-agnostic concepts carrying 87{,}518
paragraph--concept edges typed by relation and 55{,}872
concept--concept edges, in which canonical English labels act as a
string-equality key that links Bengali and English passages with no
parallel data; (iii) a bilingual alias inventory of 60{,}850 surface
forms (30{,}053 English, 30{,}797 Bengali) mapping each concept to its
realizations in both languages; and (iv) a human-annotated evaluation set
of 200 paragraph--concept edges judged independently by three annotators,
released with all per-annotator judgments. We document the construction
pipeline, which needs one extraction pass by a mid-sized instruction-tuned
LLM and rebuilds incrementally in 10--15 minutes after source edits, and
report three evaluations: known-item cross-lingual retrieval over 194
automatically verified rendered lecture pairs (Recall@10 0.86 in both
directions); a 30-question audit of citation integrity and modern-question
bridging; and the human study, which places concept-extraction precision
at 0.60 under strict two-annotator consensus (Cohen's $\kappa = 0.61$).
The study also shows the extractor's confidence weight is calibrated ---
restricting to weight $\geq 0.8$ raises precision to 0.71 while retaining
98\% of concept-bearing paragraphs --- and that precision is markedly
lower in Bengali than English (0.54 vs 0.68), locating the resource's
weakness in exactly the half that cross-lingual access depends on. The
design assumes nothing Vivekananda-specific and transfers to other
multilingual classical corpora.}


\maketitle

\section{Introduction}

Large digitized editions of classical religious and philosophical
literature are now widely available, but access to them remains largely
lexical: readers search for words, while what they hold are questions.
The gap is widest for corpora like the works of Swami Vivekananda, a
central figure of modern Vedanta whose collected English works span nine
volumes and whose Bengali writings and recorded conversations fill ten
more. The two collections overlap thematically but are \emph{not}
translations of one another; a reader fluent in only one language has no
lexical route into roughly half of the material.

A second, subtler gap is \emph{temporal}. Contemporary readers bring
contemporary questions --- about smartphone addiction, social-media envy,
or career anxiety --- to an author who died in 1902. The vocabulary of
such questions simply does not occur in the corpus, so even strong dense
retrievers underperform; yet the corpus deals extensively with the
underlying phenomena under names such as \emph{attachment},
\emph{control of the mind}, and \emph{habit}. Bridging this gap is not a
retrieval nicety but the central requirement for making a
nineteenth-century corpus useful to twenty-first-century readers.

Finally, any system that \emph{generates} answers about revered texts
must confront hallucination \cite{ji2023survey}. Attributing invented
statements to a religious figure is a materially worse failure mode than
a wrong answer in open-domain QA, so verifiable grounding --- every claim
traceable to a specific passage a reader can open --- is a hard
constraint rather than a desirable extra.

This paper contributes a \emph{resource} built around these three
requirements, together with the open pipeline that produces it and the
evaluations that characterize its quality.

\paragraph{The resource.} Four layers are released
(Section~\ref{sec:resource} documents formats and distribution):

\begin{enumerate}[leftmargin=1.4em]
  \item \textbf{A structured bilingual parse.} Both published editions are
  recovered into a volume\,$\to$\,chapter\,$\to$\,paragraph\,$\to$\,sentence
  hierarchy --- 1{,}991 chapters, 32{,}694 paragraphs, 168{,}842
  sentences --- with the HTML anchor of every paragraph retained, so any
  unit in the resource deep-links to its exact location in the public
  online editions (Section~\ref{sec:corpus}).
  \item \textbf{A cross-lingual concept graph.} An instruction-tuned LLM
  reads every paragraph of both corpora and emits concepts with
  \emph{canonical English, lowercase, hyphenated labels}. Because the
  label space is shared, a Bengali paragraph about \emph{vairagya} and an
  English lecture on renunciation attach to the same node; string
  equality on labels yields cross-lingual linking without any parallel
  data. The graph carries 8{,}362 concepts, 87{,}518 relation-typed
  paragraph--concept edges, and 55{,}872 concept--concept edges
  (Section~\ref{sec:graph}).
  \item \textbf{A bilingual alias inventory.} 60{,}850 surface forms
  (30{,}053 English, 30{,}797 Bengali) record how each concept is actually
  realized in each language --- usable independently of the rest of the
  resource as a concept-level bilingual lexicon for Vedantic vocabulary
  (Section~\ref{sec:graph}).
  \item \textbf{A human-annotated evaluation set.} 200 paragraph--concept
  edges sampled uniformly at random, each judged independently by three
  annotators, released with all per-annotator judgments so that alternative
  aggregations and agreement analyses can be recomputed
  (Section~\ref{sec:eval-precision}).
\end{enumerate}

\paragraph{The pipeline and its evaluation.} Around the resource we
document and release: multi-granular fused retrieval, which matches
queries against sentence-, paragraph- and chapter-level indexes in both
languages \emph{and} against the concept graph, fusing candidates with
weighted reciprocal rank fusion \cite{cormack2009rrf} and reranking with a
multilingual cross-encoder (Section~\ref{sec:retrieval}); temporal
concept bridging, which maps modern questions onto the concept vocabulary
that verifiably exists in the graph and surfaces the mapping as an
inspectable ``bridge'' (Section~\ref{sec:qa}); citation-grounded
generation, in which every claim carries a citation rewritten into a deep
link to the exact source paragraph and unsupported markers are deleted
(Section~\ref{sec:qa}); and an economical build --- one GPU-day for the
full corpus, 10--15 minutes for an incremental rebuild after a source
edit, which is what makes the resource maintainable rather than frozen
(Section~\ref{sec:build}).

\section{Related Work}

\textbf{Dense retrieval and RAG.} Dense passage retrieval
\cite{karpukhin2020dpr} and retrieval-augmented generation
\cite{lewis2020rag,gao2023ragsurvey} are the standard architecture for
grounded QA. We adopt the general recipe but depart in two ways: the
retrieval layer is multi-granular and graph-augmented, and generation is
constrained to a citation discipline enforced by post-processing rather
than trust in the model.

\textbf{Graph-augmented retrieval.} GraphRAG \cite{edge2024graphrag}
builds an entity graph over a corpus with an LLM and uses community
summaries for query-focused summarization. \system{} shares the
LLM-builds-a-graph idea but uses the graph differently: nodes are
\emph{concepts} with language-agnostic canonical labels, and the graph's
primary role is cross-lingual bridging and query-time expansion into two
non-parallel corpora, rather than corpus summarization.

\textbf{Multilingual embeddings.} We build on BGE-M3
\cite{chen2024bgem3}, whose single embedding space covers both English
and Bengali and supports inputs long enough for chapter-level
representation, and on its companion multilingual cross-encoder reranker.
Sentence-level transformer embeddings follow the general approach of
\cite{reimers2019sbert}.

\textbf{QA over religious corpora.} Verse-based test collections and QA
systems exist for the Qur'an \cite{malhas2020ayatec} and other
scriptures; these typically operate in a single language over a closed
canon. To our knowledge, \system{} is the first system to combine
non-parallel bilingual retrieval, an LLM-built concept layer, and
explicit modern-question bridging for a classical corpus.

\textbf{Positioning as a resource.} Existing cultural-heritage language
resources of this kind fall into three groups, and this resource sits
between them. Digitized \emph{editions} (page images, TEI transcriptions)
preserve structure and provenance but offer no semantic access layer.
\emph{Aligned bilingual corpora} support cross-lingual work but presuppose
translation pairs, which do not exist here and cannot be manufactured
without distorting the material: the Bengali collection is not a
translation of the English one. \emph{Knowledge graphs} over cultural
heritage are typically entity-centric --- people, places, works --- and
built by linking to external authority files, which for a body of
philosophical prose captures the proper nouns and misses the argument.
The layer released here is instead \emph{concept}-centric, built from
the text rather than from an authority file, and language-agnostic by
construction, which is what allows non-parallel corpora to be linked at
all. For Bengali specifically, where resources remain
comparatively scarce, the alias inventory and the annotated edge sample
also constitute reusable artefacts in their own right.

\section{Corpus and Preprocessing}
\label{sec:corpus}

The English source is the nine-volume \emph{Complete Works of Swami
Vivekananda} (lectures, letters, recorded conversations, and writings;
$\sim$8.1 million characters). The Bengali source is the ten-volume
\emph{Vani o Rachana} ($\sim$6.9 million characters), which contains
original Bengali writings and conversations as well as independent
Bengali renderings of some English lectures; no paragraph-level alignment
between the two exists, and we do not attempt to create one.

Both corpora are parsed from their published HTML editions with
structure-aware parsers that recover the volume $\to$ (section) $\to$
chapter $\to$ paragraph hierarchy and preserve both the chapter and the
per-paragraph HTML anchor identifiers, which later enable paragraph-precise
deep-linking of citations back into the publicly hosted editions. Sentences are split with language-appropriate
rules (NLTK \cite{bird2004nltk} for English; a rule-based splitter
honouring the \emph{danda} and Bengali punctuation for Bengali). Bengali
text is NFC-normalized throughout; a single normalization function is the
sole source of truth for text keys across the pipeline, which the
incremental-rebuild machinery (Section~\ref{sec:build}) depends on.
Table~\ref{tab:corpus} summarizes the corpus.

\begin{table}[t]
\centering
\caption{Corpus statistics after parsing.}
\label{tab:corpus}
\begin{tabular}{lrrr}
\toprule
 & English & Bengali & Total\\
\midrule
Volumes (published)  & 9      & 10     & 19\\
Chapters    & 1{,}452 & 539   & 1{,}991\\
Paragraphs  & 18{,}239 & 14{,}455 & 32{,}694\\
Sentences   & 88{,}150 & 80{,}692 & 168{,}842\\
Characters  & 8.1M   & 6.9M   & 15.0M\\
\bottomrule
\end{tabular}
\end{table}

\section{The Concept Graph}
\label{sec:graph}

\subsection{LLM concept extraction}

An instruction-tuned LLM (Qwen2.5-14B-Instruct \cite{qwen25}, served with
vLLM \cite{kwon2023vllm}) reads every paragraph in both languages and
emits structured JSON containing (i) a one-line English summary, (ii) a
set of \emph{concepts}, and (iii) a set of named \emph{entities} typed as
person, place, text, or deity. The extraction prompt imposes the single
most important invariant of the system: \textbf{concept labels are
canonical English, lowercase, and hyphenated} regardless of the paragraph
language. The canonical label thus acts as a cross-lingual key: the
Bengali surface form and the English surface form of a concept are stored
as \emph{aliases} (60{,}838 aliases; 30{,}051 English, 30{,}787 Bengali),
while identity lives in the label itself. Each paragraph--concept edge
carries a relation (\emph{discusses}, \emph{exemplifies}, or
\emph{contrasts}) and a weight.

Near-duplicate concepts produced by paraphrase (e.g., spelling variants)
are merged by embedding the labels and clustering above a conservative
cosine threshold (0.92). Two kinds of concept--concept edges are then
built: \emph{similar} edges (label-embedding cosine $\geq$ 0.78, top-12
neighbours per node) and \emph{co-occurs} edges (concepts extracted from
the same paragraph at least twice corpus-wide).
Table~\ref{tab:graph} summarizes the resulting graph;
Table~\ref{tab:topconcepts} lists the most frequent concepts, which
align well with the corpus's known thematic centres.

\begin{table}[t]
\centering
\caption{Concept-graph statistics.}
\label{tab:graph}
\begin{tabular}{lr}
\toprule
Concepts (canonical labels) & 8{,}362\\
Concept aliases (EN / BN) & 30{,}053 / 30{,}797\\
Entities (person / other / place / text / deity) & 3{,}688 / 2{,}372 / 1{,}275 / 846 / 492\\
Paragraph$\to$concept edges & 87{,}518\\
\quad discusses / exemplifies / contrasts & 75{,}403 / 8{,}735 / 3{,}380\\
Concept$\leftrightarrow$concept edges & 55{,}872\\
\quad similar / co-occurs & 29{,}482 / 26{,}390\\
Paragraphs with $\geq$1 concept & 23{,}363 (71.5\%)\\
\quad retained at extractor weight $\geq$ 0.8 & 22{,}914 (98.1\%)\\
\bottomrule
\end{tabular}
\end{table}

\begin{table}[t]
\centering
\caption{Most frequent concepts by paragraph mentions.}
\label{tab:topconcepts}
\begin{tabular}{lr@{\qquad}lr}
\toprule
Concept & Mentions & Concept & Mentions\\
\midrule
self-realization      & 4{,}521 & renunciation & 2{,}553\\
knowledge-realization & 3{,}858 & liberation   & 2{,}528\\
devotion              & 3{,}179 & patience     & 1{,}787\\
maya                  & 2{,}684 & duty         & 1{,}437\\
non-attachment        & 2{,}669 & karma        & 1{,}334\\
\bottomrule
\end{tabular}
\end{table}

The 71.5\% paragraph coverage is deliberate rather than a deficiency:
short fragments (single-word replies in recorded conversations, headings,
brief exclamations) legitimately carry no concept content, and forcing
extraction on them degrades graph precision.

Coverage, however, is not correctness. A two-annotator study of the
edges themselves (Section~\ref{sec:eval-precision}) puts their
strict-consensus precision at 0.60, rising to 0.71 on the
high-confidence subgraph and falling to 0.54 on Bengali material; the
graph should be read as a broad, imperfect expansion layer rather than a
curated index, and everything we claim for it below is calibrated to
that.

\subsection{Economical construction and incremental rebuilds}
\label{sec:build}

Concept extraction is the only expensive stage ($\sim$3.5 hours for the
full corpus on a single 80\,GB-class GPU); everything else (parsing,
embedding, graph assembly) takes minutes. To make the corpus
\emph{editable} --- OCR fixes and structural cleanups are routine in
digitized classical texts --- the pipeline snapshots all extraction
outputs keyed by (NFC-normalized paragraph text, language, extractor
version) before any re-parse, and restores them onto the fresh paragraph
rows afterwards. Only genuinely new or changed paragraphs reach the LLM.
An incremental rebuild after a content edit completes in 10--15 minutes,
which in practice is the difference between a maintained resource and a
frozen one.

\section{The Released Resource}
\label{sec:resource}

\subsection{Data model and distribution format}

The resource is distributed as a single normalized SQLite database
together with the dense indexes built from it. SQLite was chosen
deliberately over a graph database or a bespoke format: it needs no
server, is readable by standard tooling in every major language, travels
as one file, and keeps the corpus, the concept layer, and their
provenance in one queryable place. Table~\ref{tab:resource} lists the
released layers and their sizes.

\begin{table}[t]
\centering
\caption{Layers of the released resource. All relational layers live in a
single 232\,MB SQLite file; the dense indexes add 868\,MB.}
\label{tab:resource}
\begin{tabular}{llr}
\toprule
Layer & Relation & Units\\
\midrule
\multirow{5}{*}{Structural parse}
  & \texttt{books}            & 2\\
  & \texttt{volumes}          & 10\\
  & \texttt{chapters}         & 1{,}991\\
  & \texttt{paragraphs}       & 32{,}694\\
  & \texttt{sentences}        & 168{,}842\\
\midrule
\multirow{4}{*}{Concept layer}
  & \texttt{concepts}         & 8{,}362\\
  & \texttt{concept\_aliases} & 60{,}850\\
  & \texttt{para\_concept}    & 87{,}518\\
  & \texttt{concept\_edges}   & 55{,}872\\
\midrule
\multirow{2}{*}{Entity layer}
  & \texttt{entities}         & 8{,}673\\
  & \texttt{para\_entity}     & 41{,}937\\
\midrule
Dense indexes
  & 7 FAISS indexes           & 211{,}889 vectors\\
\midrule
Human annotations
  & 200 items $\times$ 3 annotators & 600 judgments\\
\bottomrule
\end{tabular}
\end{table}

Every unit carries its provenance. A paragraph row records its chapter,
its ordinal position, its language, and the HTML anchor
(\texttt{para\_id\_html}) of the paragraph in the published online
edition, so any row in any layer can be resolved to a URL that opens the
exact passage a reader needs to check. Each paragraph--concept edge
carries the relation the extractor assigned (\emph{discusses},
\emph{exemplifies}, \emph{contrasts}) and its confidence weight, which
Section~\ref{sec:eval-precision} shows to be calibrated and therefore
usable as a quality filter by downstream reusers.

\subsection{What each layer offers independently}

The layers are useful separately, not only as inputs to the retrieval
stack described below:

\begin{itemize}[leftmargin=1.4em]
  \item The \textbf{structural parse} is, to our knowledge, the first
  machine-readable sentence-segmented edition of either corpus with
  stable per-paragraph identifiers, and is directly reusable for
  stylometry, citation studies, or translation research.
  \item The \textbf{concept graph} supplies cross-lingual links between
  two corpora for which no parallel data exists, and can be used as
  distant supervision for training bilingual retrievers.
  \item The \textbf{alias inventory} --- 30{,}053 English and 30{,}797
  Bengali surface forms grouped under 8{,}362 canonical concepts --- is
  usable on its own as a concept-level bilingual lexicon of Vedantic
  vocabulary, a domain poorly covered by general-purpose bilingual
  dictionaries.
  \item The \textbf{annotation set} ships with every individual
  annotator's judgments rather than an aggregated gold label, so that
  reusers can apply their own aggregation, recompute agreement, or study
  annotator disagreement on semantic-relatedness judgments in a
  bilingual setting.
\end{itemize}

\subsection{Availability and licensing}
\label{sec:availability}

The pipeline source code is released under the MIT licence at
\url{https://github.com/tamalrkm/viveka-insight}. The derived data layers --- the structural parse,
the concept graph, the alias inventory, the entity layer, and the
annotation set --- are released under CC~BY~4.0 and archived with a
persistent identifier at \texttt{10.5281/zenodo.21669973}. The dense indexes are
rebuildable from the database and the released pipeline in
Section~\ref{sec:build}'s stated budget, and are provided for convenience
rather than as the primary artefact.

The underlying texts require a distinction. Vivekananda died in 1902 and
his own words are in the public domain, but the modern published editions
from which we parse --- and, in particular, the Bengali renderings of
English lectures, which are the work of later translators --- carry
editorial and translational rights held by their publishers. We therefore
do \emph{not} redistribute the source HTML. What we release are the
derived layers plus the anchors needed to align them against the publicly
hosted editions, so the resource is fully reproducible by a user who
obtains the source texts through the publishers' own channels, and the
pipeline regenerates every layer from them.

\section{Multi-Granular Fused Retrieval}
\label{sec:retrieval}

\begin{figure}[t]
\centering
\definecolor{figgrey}{HTML}{BFBFBF}
\definecolor{figgreen}{HTML}{D9EAD3}
\definecolor{figblue}{HTML}{CFE2F3}
\definecolor{figyellow}{HTML}{FCE7C3}
\definecolor{figpink}{HTML}{F7D2D5}
\resizebox{\textwidth}{!}{%
\begin{tikzpicture}[
  font=\small,
  box/.style={align=center, inner sep=7pt, minimum height=13mm},
  arr/.style={-{Stealth[length=3mm,width=2.6mm]}, line width=1.1pt,
              draw=gray!75, rounded corners=6pt}
]
\node[box, fill=figgrey, minimum width=26mm] (q) at (0,0)
  {Query\\(EN or BN)};
\node[box, fill=figgreen, minimum width=52mm, minimum height=17mm]
  (vec) at (5.4,2.1) {Sentence/Paragraph/Chapter\\
  FAISS indexes $\times$ \{EN, BN\}};
\node[box, fill=figgreen, minimum width=52mm, minimum height=17mm]
  (con) at (5.4,-2.1) {Concept Index +\\
  1-hop graph-walk $\rightarrow$\\ Linked Paragraphs};
\node[box, fill=figblue, minimum width=30mm] (rrf) at (10.6,0)
  {Weighted RRF\\Fusion};
\node[box, fill=figyellow, minimum width=34mm] (rer) at (14.4,0)
  {Cross Encoder\\rerank (Top 60)};
\node[box, fill=figpink, minimum width=46mm] (out) at (18.9,0)
  {Ranked Passages +\\\emph{via-concept} explanations};
\draw[arr] (q.east)   -- ++(0.45,0) |- (vec.west);
\draw[arr] (q.east)   -- ++(0.45,0) |- (con.west);
\draw[arr] (vec.east) -- ++(0.45,0) |- (rrf.west);
\draw[arr] (con.east) -- ++(0.45,0) |- (rrf.west);
\draw[arr] (rrf.east) -- (rer.west);
\draw[arr] (rer.east) -- (out.west);
\end{tikzpicture}%
}
\caption{Retrieval architecture. Path A (top) matches the query against
three granularities per language; Path B (bottom) routes it through the
concept graph. Both feed a single fusion and reranking stage.}
\label{fig:arch}
\end{figure}
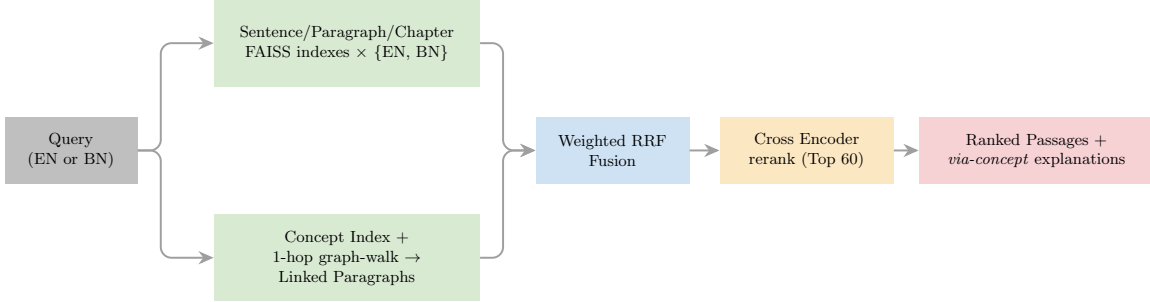

All text units --- 168k sentences, 33k paragraphs, 2k chapters, and the
8.4k concept labels --- are embedded into the same 1024-dimensional
space with BGE-M3 \cite{chen2024bgem3} and stored in exact
inner-product FAISS indexes \cite{johnson2019faiss} (about 1.1\,GB in
total including metadata, small enough that approximate indexing is
unnecessary). Retrieval runs two paths per language
(Figure~\ref{fig:arch}):

\textbf{Path A (direct).} The query is matched against the sentence
(top-30), paragraph (top-20), and chapter (top-10) indexes. Sentence and
chapter hits are mapped to their enclosing/leading paragraphs, since the
paragraph is the unit surfaced to users; the granularity that found each
hit is retained and displayed.

\textbf{Path B (concept-mediated).} The query is matched against the
concept-label index (top-12 anchors), the anchors are expanded one hop
along \emph{similar} and \emph{co-occurs} edges with weight decay, and
each activated concept contributes its top linked paragraphs. Because
concept nodes are language-agnostic, an English query activates Bengali
paragraphs (and vice versa) with no translation step. Path B also yields
an explanation --- the concepts through which each paragraph was reached
--- which the interface displays as ``via concepts'' badges.

Candidates from the four rank lists (three granularities + concept path)
are fused with weighted reciprocal rank fusion \cite{cormack2009rrf},
with the concept path weighted slightly above the direct paths (1.2 vs.\
1.0) and chapters below (0.7). The top 60 fused candidates are reranked
by the BGE-reranker-v2-m3 cross-encoder, and the top $k$ per language are
returned with full provenance (volume, section, chapter, paragraph index,
and a deep link to the chapter anchor in the hosted edition).

\section{``Ask Vivekananda'': Bridged, Citation-Grounded QA}
\label{sec:qa}

The QA layer turns retrieval into direct answers under two hard
constraints: modern questions must reach era-appropriate material, and
every generated claim must be verifiable. The pipeline has three stages.

\textbf{Stage 1: analyze and bridge.} The question's 15 nearest concept
labels are fetched from the concept index. A single LLM call receives the
question \emph{and this list} and returns (i) 2--4 reformulated
``timeless'' search queries free of modern-only vocabulary, (ii) up to 6
concept labels \emph{copied from the offered list}, and (iii) a one-line
\emph{bridge note} explaining the modern-to-timeless mapping. Restricting
concept choice to labels that verifiably exist in the graph (validated
again against the database after generation) prevents the bridge itself
from hallucinating; if JSON parsing fails, the system degrades to
retrieval with the raw question. The question's language is detected
deterministically from its script, never delegated to the LLM.

\textbf{Stage 2: retrieve.} The original question and each reformulation
are run through the full retrieval stack of
Section~\ref{sec:retrieval}; results are merged, deduplicated per
(language, paragraph), and the pooled candidates are reranked once
against the \emph{original} question. The top 10 passages become the
numbered sources.

\textbf{Stage 3: generate and ground.} A locally hosted
Qwen2.5-7B-Instruct \cite{qwen25} (served via Transformers
\cite{wolf2020transformers}; any OpenAI-compatible endpoint can be
substituted by configuration) receives the numbered sources ---
location-labelled, language-tagged, and trimmed to a window centred on
each passage's best-matching sentence --- and composes a scholarly
third-person answer in the language of the question. The system prompt
forbids claims not supported by the sources and requires a bracketed
citation after every substantive statement. Post-processing then rewrites
each citation marker $[n]$ into a hyperlink into the published online
edition and \emph{deletes} any marker whose index does not correspond to a
retrieved source, so a fabricated citation cannot survive to the reader.
The published editions carry a stable HTML anchor on every paragraph, so
each citation deep-links to the \emph{exact cited paragraph} (falling back
to the chapter anchor where a paragraph anchor is absent) --- the reader
lands on the sentence in question, not merely the top of a chapter. The
interface additionally shows the bridge note,
the reformulated queries, the chosen concepts, and, for every source, the
matched sentence and the full passage --- the entire evidential chain is
one click deep.

\section{System Characteristics}
\label{sec:stats}

\begin{table}[t]
\centering
\caption{Serving-time characteristics (single NVIDIA H200; the system
also runs CPU-only with seconds-level search latency). Model download
sizes excluded.}
\label{tab:perf}
\begin{tabular}{lr}
\toprule
Search, with cross-encoder rerank (mean of 5 queries) & 0.28\,s\\
Search, fusion only & 0.08\,s\\
QA end-to-end (bridge + retrieve + generate) & 6--9\,s\\
Index footprint (7 FAISS indexes + SQLite metadata/graph) & 1.1\,GB\\
Full pipeline build (one GPU) & $<$ 4\,h\\
Incremental rebuild after a source edit & 10--15\,min\\
\bottomrule
\end{tabular}
\end{table}

The stack is deliberately conservative: exact FAISS search, SQLite for
all metadata and graph storage, and a Streamlit interface. There are no
external service dependencies at query time; the system runs entirely
on-premises, which matters both for cost and for institutions that
cannot ship religious-community data to third-party APIs.
Table~\ref{tab:perf} reports measured characteristics. A test suite of
74 CPU-only tests covers the parsers, schema, extraction-output
parsing, fusion, bridging fallbacks, prompt budgeting, citation
linkification, and the concept-graph view.

\section{Evaluation}
\label{sec:eval}

We report three studies: two automatic ones that require no manual
annotation, and a human study of the concept layer. All are reproducible
with evaluation scripts shipped in the
repository.\footnote{\texttt{scripts/eval\_crosslingual.py},
\texttt{scripts/eval\_qa\_integrity.py}, and
\texttt{scripts/eval\_concept\_precision.py}; the raw outputs quoted
here, including the per-annotator judgment files, are included under
\texttt{docs/paper/eval/}.}

\subsection{Known-item cross-lingual retrieval}
\label{sec:eval-knownitem}

The \emph{Vani o Rachana} contains independent Bengali renderings of many
English lectures, which yields chapter-level cross-lingual relevance
judgments at near-zero annotation cost. We identified candidate pairs as
mutual-best matches between the chapter-level embedding indexes (cosine
$\geq 0.70$; 285 candidates), then verified each candidate with the local
LLM shown both chapters' titles and opening paragraphs, retaining
$N = 194$ verified rendered pairs. For each pair, the source chapter's
title and opening paragraph (truncated to 600 characters) is the query,
retrieval is restricted to the \emph{other} language, and the paired
chapter is the sole gold item; a retrieved paragraph counts if it belongs
to the gold chapter. We report Recall@10 and MRR over the top 10 returned
paragraphs in both directions (Table~\ref{tab:knownitem}).

One caveat: pair discovery uses the same embedder as Path A (albeit on
whole-chapter vectors, while queries are titles plus opening paragraphs),
so absolute Path-A numbers may be slightly optimistic; comparisons
\emph{across} configurations are unaffected.

\begin{table}[t]
\centering
\caption{Known-item cross-lingual retrieval over $N=194$ LLM-verified
rendered lecture pairs (EN$\to$BN / BN$\to$EN).}
\label{tab:knownitem}
\begin{tabular}{lcc}
\toprule
Configuration & Recall@10 & MRR\\
\midrule
Path A (direct dense)        & 0.88 / 0.82 & 0.74 / 0.65\\
Path B (concept-mediated)    & 0.14 / 0.12 & 0.07 / 0.04\\
Fused (weighted RRF)         & 0.88 / 0.82 & 0.67 / 0.62\\
Fused + cross-encoder rerank & 0.86 / 0.86 & 0.67 / 0.70\\
\bottomrule
\end{tabular}
\end{table}

Three observations. First, the shared multilingual embedding space alone
is a strong cross-lingual known-item retriever: a query built from an
English lecture finds its Bengali rendering among 14{,}455 Bengali
paragraphs at Recall@10 of 0.88. Second, the concept path is
\emph{not} a known-item retriever and is not meant to be one --- it
expands a query into thematically related material across the whole
graph, which is precisely what known-item metrics penalize; its value
shows in query bridging and in the explanatory ``via concepts'' chains,
not in this table. Third, fusion preserves Path A's recall while adding
the concept candidates, and the cross-encoder markedly improves the
harder BN$\to$EN direction (MRR 0.62$\to$0.70, Recall@10
0.82$\to$0.86) at a small cost in EN$\to$BN, yielding the most balanced
configuration --- the production default.

\subsection{Citation integrity and bridge reliability}
\label{sec:eval-integrity}

We ran the full QA pipeline over a fixed set of 30 questions --- 20
English and 10 Bengali; 15 phrased in deliberately modern vocabulary
(smartphone addiction, social-media envy, job-loss anxiety, burnout) and
15 in timeless vocabulary (fear, concentration, duty, anger) --- and
logged the pipeline's internal decisions (Table~\ref{tab:integrity}).

\begin{table}[t]
\centering
\caption{Citation-integrity and bridging statistics over 30 questions
(20 EN / 10 BN; 15 modern / 15 timeless).}
\label{tab:integrity}
\begin{tabular}{lr}
\toprule
Questions answered (non-empty retrieval) & 30 / 30\\
Stage-1 JSON fallback rate & 0\%\\
Bridge note produced: modern questions & 14 / 15 (93\%)\\
Bridge note produced: timeless questions & 2 / 15 (13\%)\\
Citation markers per answer sentence & 0.47\\
Citation markers emitted (total) & 136\\
Fabricated markers removed by validator & 0 / 136 (0\%)\\
Cross-lingual share of top-10 evidence (bridged) & 32\%\\
Cross-lingual share of top-10 evidence (raw question) & 40\%\\
\bottomrule
\end{tabular}
\end{table}

The bridging stage discriminates almost perfectly between question types:
it produced a bridge note for 93\% of modern questions and only 13\% of
timeless ones, with a 0\% structured-output failure rate. Generated
answers cite densely (one marker per two sentences on average), and in
this run the 7B model emitted \emph{no} citation marker pointing outside
the retrieved source list --- the validator, which deletes such markers,
acted as an idle guardrail rather than an active repair mechanism.

The bridge ablation produced a finding we did not anticipate: the
cross-lingual share of the evidence pool is essentially unchanged by
bridging (32\% bridged vs.\ 40\% raw; 47\% vs.\ 45\% on the modern
subset). Cross-linguality is supplied by the shared embedding space and
the concept graph themselves, not by the reformulation step. The bridge's
measured contributions are therefore selective activation (firing when
and only when the vocabulary gap exists), the era-appropriate
reformulations, and the user-facing explanation --- not an increase in
cross-lingual recall, which the underlying retrieval already provides.

\subsection{Concept-extraction precision: a human study}
\label{sec:eval-precision}

The concept layer is the system's central claim, and it is produced
entirely by an LLM. We therefore ran a human study of its precision.

\textbf{Protocol.} From all 87{,}518 paragraph--concept edges we drew a
uniform random sample of $N=200$ (seed 13). Each item presents one
paragraph, one attached concept label, and the extracted relation.
Annotators judged each item as \emph{correct} (the concept is a correct
reading of the paragraph \emph{at the stated relation}) or
\emph{incorrect} (wrong, merely tangential, or right concept but wrong
relation). Judgments were collected through a small purpose-built web
interface shipped with the repository, which saves after every click and
lets an annotator resume; no annotator saw another's judgments or the
extractor's confidence weight.

\textbf{Annotators.} Three annotators, all familiar with the corpus, each
judged all 200 items independently. Annotators~A and~B are fluent in
both English and Bengali; annotator~C is proficient principally in
Bengali. Since 85 of the 200 sampled items are English, we use C's
judgments only for the Bengali subsample, where that proficiency applies,
and base the whole-sample figures on A and B. All three judgment files
ship with the repository.\footnote{\texttt{docs/paper/eval/annotations/};
the whole-sample figures below are reproduced with
\texttt{eval\_concept\_precision.py compute --users annotator\_a annotator\_b}, and any
other pairing can be recomputed from the same files.}

That scoping is a competence requirement fixed by the task, and the
measured reliability tracks it closely: against~A, annotator~C reaches
$\kappa = 0.49$ on the Bengali items but only $\kappa = 0.14$ --- barely
above chance --- on the English ones. Restricting C to Bengali is
therefore not a way of setting aside an inconvenient judge but of using
each annotator where they are reliable; we report the Bengali
three-annotator panel in full below, and it does not flatter the system.

\textbf{Disclosure.} Annotator~B is an author of this paper; annotators~A
and~C are not. B is marginally the more lenient of A and B (0.705 vs
0.665), so we treat the strict consensus figure --- not B's --- as the
headline number.

\begin{table}[t]
\centering
\caption{Concept-extraction precision over $N=200$ randomly sampled
paragraph--concept edges, annotators A and B. Intervals are Wilson 95\%
(bootstrap for $\kappa$).}
\label{tab:precision}
\begin{tabular}{lcc}
\toprule
Measure & Value & 95\% CI\\
\midrule
Precision, annotator A (non-author)   & 0.665 & [0.597, 0.727]\\
Precision, annotator B (author)       & 0.705 & [0.638, 0.764]\\
\textbf{Precision, consensus} (both correct) & \textbf{0.600} & [0.531, 0.665]\\
Precision, union (either correct)     & 0.770 & [0.707, 0.823]\\
\midrule
Raw agreement                         & 0.830 & ---\\
Cohen's $\kappa$ \cite{cohen1960kappa} & 0.607 & [0.482, 0.720]\\
Gwet's AC$_1$ \cite{gwet2008ac1}       & 0.701 & ---\\
\bottomrule
\end{tabular}
\end{table}

\textbf{Precision is moderate; agreement is substantial.} Consensus
precision is 0.60 --- three in five sampled edges are endorsed by both
judges --- and Cohen's $\kappa$ of 0.61 falls in the ``substantial'' band
on the conventional scale \cite{landis1977kappa}. The residual
disagreement is not systematic: 21 items go one way and 13 the other,
which a McNemar exact test does not distinguish from chance
($p = 0.23$). The two are applying the same rubric with ordinary noise,
which is what an inter-annotator agreement study is supposed to establish
before its precision figure is worth quoting.

\textbf{The extractor's confidence weight is calibrated.} Each
paragraph--concept edge carries a weight the extraction LLM assigns
itself. It predicts human judgment (Spearman $\rho = 0.19$ against
consensus), and thresholding on it works
(Table~\ref{tab:precision-weight}): consensus precision rises from 0.54
below $0.8$ to 0.71 at $0.8$ and above (Fisher exact $p = 0.023$).

\begin{table}[t]
\centering
\caption{Precision by extractor confidence weight. The high-confidence
subgraph is 35.9\% of all edges but still covers 98.1\% of the
paragraphs that carry any concept.}
\label{tab:precision-weight}
\begin{tabular}{lccc}
\toprule
 & Annot.\ A & Annot.\ B & Consensus\\
\midrule
weight $\geq 0.8$ \ ($n=69$)  & 0.783 & 0.812 & \textbf{0.710}\\
weight $< 0.8$ \ ($n=131$)    & 0.603 & 0.649 & 0.542\\
\bottomrule
\end{tabular}
\end{table}

The pruning is nearly free in coverage terms. Applied to the full graph,
weight $\geq 0.8$ retains 31{,}445 of 87{,}518 paragraph--concept edges
(35.9\%) and 3{,}709 of 8{,}362 concepts (44.4\%) --- but
22{,}914 of the 23{,}363 paragraphs that carry any concept
(\textbf{98.1\%}) keep at least one. The low-confidence mass is
redundant attachment to already-covered paragraphs, not the sole
handle on distinct material. A deployment that needs the graph to be
\emph{right} rather than \emph{broad} should therefore threshold at
$0.8$; we report the unpruned graph throughout this paper because Path~B
is used for recall-oriented expansion, where the cost of a wrong concept
is a wasted glance rather than a false statement.

\textbf{Extraction is weaker in Bengali.} Consensus precision over A and
B is 0.68 on English items and 0.54 on Bengali (Fisher exact
$p = 0.043$). The Bengali subsample ($n = 115$) is the one part of the
study carrying a third independent judgment, and the fuller panel
confirms the gap rather than closing it: across A, B and~C the
three-rater Fleiss $\kappa$ is 0.49 and majority-of-three precision is
0.59, still well below the 0.68 measured on English. Adding a
Bengali-proficient annotator thus corroborates the finding with the
language's strongest available judgement.

This is the study's one clearly actionable quality finding beyond the
threshold: the extraction LLM is a predominantly English-trained model
reading Bengali prose, and the concept layer inherits that asymmetry.
Since cross-lingual bridging is the system's central claim, the weaker
half of the graph is precisely the half doing the work of making Bengali
material reachable from English queries --- a Bengali-strong extractor,
or a second extraction pass over Bengali paragraphs alone, is the most
valuable single upgrade available to the pipeline.

\textbf{Relations.} The dominant \emph{discusses} (87.5\% of the sample)
reaches consensus precision 0.62, while the two rare relations are worse
(\emph{exemplifies} 0.47, $n=17$; \emph{contrasts} 0.50, $n=8$) --- on
samples too small to support a strong claim, but consistent with the
intuition that a directional judgment is harder for the extractor than a
topical one.

\textbf{On the rubric.} The binary judgment conflates two readings of
``the concept describes the paragraph'': whether the paragraph \emph{is
about} the concept, and whether the concept is merely \emph{present} in
it. For a S\=utra gloss such as ``The success of Yogis differs according
as the means they adopt are mild, medium, or intense'' tagged
\texttt{success}, or a Bengali travel note announcing a forthcoming
Vedanta lecture tagged \texttt{vedanta}, the first reading rejects and
the second accepts. Both annotators converged on the more permissive
presence reading, so the 0.60 we report should be understood as a
precision under that reading; a study that fixed the stricter aboutness
criterion for every annotator would return a lower number. A revised
study should use a three-point scale (\emph{central} / \emph{mentioned} /
\emph{unsupported}) rather than force this distinction into a binary, and
should fix the reading in the guidelines instead of leaving it to be
resolved per annotator.

\textbf{What the errors look like.} The characteristic failure is not a
bizarre concept but a plausible one attached to a paragraph that does not
support it. Narrative, epistolary, and connective passages attract the
corpus's thematic vocabulary regardless of content: a Bengali letter
remarking that a book ``has been well received here, but the publishers
do not seem to be pushing sales'' (our translation) is tagged
\texttt{knowledge-realization}; a subscription appeal for a memorial is
tagged \texttt{liberation}. The
extractor appears to condition on the corpus's overall register as much
as on the paragraph in front of it. This is precisely the failure mode
that the confidence threshold suppresses.

\textbf{Implication for the system's claims.} The concept layer should
be understood as a \emph{recall-oriented, imperfect expansion and
explanation layer}, not a curated ontology. This is consistent with
Table~\ref{tab:knownitem}, where Path~B is a poor known-item retriever
on its own; its contribution is cross-lingual reach and the
``via concepts'' explanations, both of which degrade gracefully under
error because every retrieved passage is displayed to the reader with
its source. A wrong concept costs an irrelevant result, not a false
assertion about Vivekananda --- the citation discipline of
Section~\ref{sec:qa}, not the concept graph, is what bounds that risk.

\section{Case Studies}
\label{sec:cases}

We illustrate the bridging behaviour with unedited system outputs.

\textbf{Modern question, English.} For \emph{``How do I get rid of
mobile phone addiction?''} the analysis stage produced the bridge note
\emph{``Mobile-phone addiction is a modern form of attachment of the
senses and a habit that enslaves the mind''}, selected the corpus
concepts \texttt{excessive-attachment} and \texttt{self-control}, and
reformulated the question into queries such as \emph{``How can I
overcome excessive attachment to sense objects?''} and \emph{``What is
the remedy for habit that enslaves the mind?''}. Retrieval returned ten
passages spanning \emph{Inspired Talks}, the Raja-Yoga lectures, and ---
notably --- Bengali chapters on liberation (\emph{mukti}) and
sense-withdrawal (\emph{pratyahara}), which no lexical and no unbridged
dense query for ``mobile phone'' reaches. The generated answer opens by
restating the bridge, then prescribes non-attachment and graded practice
of sense-withdrawal, with every paragraph citing its sources.

\textbf{Modern question, second example.} For \emph{``How should one
deal with social media envy?''} the bridge note read \emph{``Social
media envy is a modern form of coveting the social status and approval
of others\ldots''} --- a mapping to covetousness and approval-seeking
that the authors did not anticipate but that the concept inventory
supported.

\textbf{Bengali question.} A Bengali question on freedom from mobile
addiction (\emph{``mobile asakti theke muktir upay ki?''}) was detected
as Bengali, bridged through the same English-labelled concepts, answered
\emph{in Bengali}, and grounded predominantly in Bengali sources while
still drawing on English lectures --- the concept layer making the
cross-lingual evidence pool available in both directions.

\section{Limitations and Ethical Considerations}
\label{sec:limitations}

\textbf{An imperfect concept layer.} The human study of
Section~\ref{sec:eval-precision} reports strict-consensus precision of
0.60 over paragraph--concept edges, and 0.54 on Bengali material. The
concept graph is a useful expansion and explanation layer, but it is not
a scholarly index and should not be cited as one. Users who need higher
precision should threshold at extractor weight $\geq 0.8$, which lifts
consensus precision to 0.71 at almost no cost in paragraph coverage.

\textbf{Limits of that study.} The whole-sample figure rests on two
annotators, one of them an author of this paper and marginally the more
lenient of the two, judging 200 items --- enough to place the precision
within roughly $\pm 0.07$, but not to support fine comparisons. A third
annotator, proficient principally in Bengali, contributes only to the
Bengali subsample; the English half therefore carries no judgement from
outside the author's immediate circle. The binary rubric conflated ``the
paragraph is about this concept'' with ``this concept appears in the
paragraph'' and left the choice to each annotator; the two whole-sample
annotators settled on the more permissive reading, so the headline figure
is the optimistic end of the plausible range --- the Bengali-proficient
annotator, who judged more strictly, would place it lower. A three-point
scale, guidelines that fix the reading, and a larger and wholly
independent annotator pool are all needed before the precision figure can
be treated as settled rather than indicative.

\textbf{Answer faithfulness is still unevaluated.} Citation integrity is
measured structurally (marker validity), not semantically: we verify that
every citation points at a retrieved passage, not that the passage
supports the claim attached to it. Known-item judgments likewise come
from rendered lecture pairs rather than topical relevance assessment. A
scholar-annotated faithfulness study of generated answers remains the
most important piece of future work.

\textbf{Generation quality floor.} The default 7B answer model
occasionally paraphrases loosely when quoting, and may present material
from a Bengali source in English without consistently marking it as
translated, despite prompt instructions. The citation mechanics bound
the damage --- the linked passage is always one click away --- but do not
eliminate it. Configurable substitution of a stronger model raises the
floor.

\textbf{Extraction bias.} The concept inventory reflects the judgement
of the extraction LLM; systematic blind spots in that model's reading of
Vedantic material would propagate to the graph. The alias tables and
per-paragraph provenance make audits feasible.

\textbf{Cultural sensitivity.} The corpus is scripture-adjacent for a
living community. We consider verifiable grounding, the visible bridge
explanation, and on-premises deployment to be ethical requirements for
this material, not merely engineering choices, and we would caution
against deploying fluent-generation systems over such corpora without
citation enforcement of comparable strictness.

\section{Conclusion}

\system{} demonstrates that a modest, fully local pipeline --- one
extraction pass by a mid-sized LLM, a shared multilingual embedding
space, and a canonical-label concept graph --- suffices to make a
bilingual, non-parallel classical corpus searchable by meaning,
answerable in either language, and honest about its evidence. The
temporal bridge turns the system's most obvious weakness (a
nineteenth-century vocabulary) into an explicit, inspectable mapping
rather than a silent retrieval failure. The architecture assumes nothing
Vivekananda-specific: any corpus with stable canonical structure and one
or more languages covered by the embedding model --- the Ramakrishna
\emph{Kathamrita} literature, Gandhi's collected works in English,
Gujarati and Hindi, or patristic corpora in Greek and translation ---
could be indexed with the same pipeline. The human study also marks the
clearest directions for the next iteration. At strict-consensus precision
of 0.60 the concept layer is broad but imperfect, and two of its
properties point at cheap gains: the extractor's own confidence weight
separates good edges from bad ones well enough that a calibrated
threshold buys accuracy almost for free, and the precision gap between
English (0.68) and Bengali (0.54) says the largest single quality deficit
is not the pipeline but the extraction model's weaker reading of Bengali
--- the very half of the corpus that cross-lingual bridging exists to
open up. A Bengali-capable extractor and community-driven correction
through the audit trail the system already exposes are therefore the
first steps; graph-embedding refinement of the concept layer and a
scholar-annotated faithfulness study of generated answers are the
further ones.

\backmatter

\bmhead{Acknowledgements}

Facilities utilized in this research were supported by the Fund for
Improvement of S\&T Infrastructure (FIST) program of the Department of
Science and Technology (DST), India [Sanction No.: SR/FST/MS-I/2022/116].

The author thanks Swami Shastravidyananda and a second volunteer
annotator, who each independently judged all 200 items of the
concept-precision study reported in Section~\ref{sec:eval-precision}.

\section*{Statements and Declarations}

\bmhead{Funding}
Facilities used in this research were supported by the Fund for
Improvement of S\&T Infrastructure (FIST) program of the Department of
Science and Technology (DST), India [Sanction No.:
SR/FST/MS-I/2022/116]. No other funding was received.

\bmhead{Competing interests}
The author declares no competing financial interests. One
non-financial interest is disclosed: the author served as one of the
three annotators in the concept-precision study of
Section~\ref{sec:eval-precision}. The study reports each annotator
separately, treats the strict two-annotator consensus rather than the
author's own more lenient judgments as the headline figure, and releases
all per-annotator judgments so that any aggregation can be recomputed
independently.

\bmhead{Ethics approval}
Not applicable. The study involved no human subjects: the three
annotators contributed expert judgments on published text as
acknowledged contributors, not as research participants, and no personal
data were collected.

\bmhead{Data availability}
The derived resource layers described in Section~\ref{sec:resource} ---
structural parse, concept graph, alias inventory, entity layer, and the
200-item annotated evaluation set with all per-annotator judgments ---
are archived under CC~BY~4.0 at \texttt{10.5281/zenodo.21669973}. The source texts of
the published editions are not redistributed; see
Section~\ref{sec:availability}.

\bmhead{Code availability}
The complete construction and serving pipeline, including the three
evaluation scripts that reproduce every figure reported here, is
released under the MIT licence at \url{https://github.com/tamalrkm/viveka-insight}.

\bmhead{Author contributions}
T.M.\ designed and implemented the system, conducted the evaluations, and
wrote the manuscript.

\end{document}